\documentclass[12pt]{article}

\usepackage{times}
\usepackage{graphicx}
\usepackage{xcolor}
\usepackage{subcaption}
\usepackage{makecell}
\usepackage{amsmath} 
\usepackage{amssymb}  
\usepackage{color,soul}
\usepackage{xfrac}
\usepackage{graphicx}
\usepackage{tikz}
\usetikzlibrary{automata, positioning}
\usetikzlibrary {decorations.pathmorphing}
\usepackage{tikz-qtree}
\usepackage{forest}
\usepackage{mathtools} 
\usepackage{mathrsfs}

\usepackage[labelfont=bf]{caption}
\usepackage{float}
\usepackage{adjustbox}
\usepackage[numbers]{natbib}

\newenvironment{sciabstract}{%
\begin{quote} \bf}
{\end{quote}}

\newcommand{\sub}[1]{_{\text{#1}}}
\newcommand{\p}[1]{\text{P}_\text{#1}}

\newcommand{\Pm}{\mathbf{\underline{P}}}
\newcommand{\Qm}{\mathbf{\underline{Q}}}
\newcommand{\Nm}{\mathbf{\underline{N}}}
\newcommand{\Eye}{\mathbf{\underline{I}}}
\newcommand{\Mzro}{\mathbf{\underline{0}}}
\newcommand{\MTF}{\text{MTF}}
\newcommand{\MTS}{\text{MTS}}
\newcommand{\MTN}{\text{MTN}}

\newcommand{\cd}{\text{\leavevmode\unskip$\cdot$\ignorespaces}}
\newcommand{\minus}{\text{\textendash}}
\newcommand{\+}{\text{\leavevmode\unskip$+$\ignorespaces}}

\title{Optimal decision making in robotic assembly and other trial-and-error tasks}

\author
{James Watson$^{1\ast}$, Nikolaus Correll$^{1}$\\
\\
\normalsize{$^{1}$Department of Computer Science, University of Colorado}\\
\normalsize{Boulder, CO, 80309 USA}\\
\\
\normalsize{$^\ast$To whom correspondence should be addressed; E-mail:  james.watson-2@colorado.edu.}
}

\date{}

\begin{document} 


\baselineskip24pt


\maketitle 

\section{Summary}

\begin{sciabstract}
We present an analytical model for when to preempt a failing robotic trial-and-error task to maximize time efficiency.

\end{sciabstract}

\section{Abstract}


\begin{sciabstract}
   Uncertainty in perception, actuation, and the environment often require multiple attempts for a robotic task to be successful. We study a class of problems providing (1) low-entropy indicators of terminal success / failure, and (2) unreliable (high-entropy) data to predict the final outcome of an ongoing task. Examples include a robot trying to connect with a charging station, parallel parking, or assembling a tightly-fitting part. The ability to restart after predicting failure early, versus simply running to failure, can significantly decrease the makespan, that is, the total time to completion, with the drawback of potentially short-cutting an otherwise successful operation. Assuming task running times to be Poisson distributed, and using a Markov Jump process to capture the dynamics of the underlying Markov Decision Process, we derive a closed form solution that predicts makespan based on the confusion matrix of the failure predictor. This allows the robot to learn failure prediction in a production environment, and only adopt a preemptive policy when it actually saves time. We demonstrate this approach using a robotic peg-in-hole assembly problem using a real robotic system. Failures are predicted by a dilated convolutional network based on force-torque data, showing an average makespan reduction from 101s to 81s ($N=120$, $p<0.05$). We posit that the proposed algorithm generalizes to any robotic behavior with an unambiguous terminal reward, with wide ranging applications on how robots can learn and improve their behaviors in the wild. 
\end{sciabstract}

\section{Introduction}

The concept of ``trial-and-error'' learning, or improving performance while doing the same task over and over again, has been extensively studied in the context of acquiring manipulation skills in both humans \cite{crossman1959theory} and robots \cite{ebert2018robustness}. Learning can increase accuracy, precision, and throughput in industrial settings. However, uncertainty in perception, actuation, and the environment makes it unlikely that random error can be entirely eliminated from task execution. For example, tasks such as peg-in-hole insertions during furniture assembly \cite{IKEAsciRob}, parallel parking \cite{paromtchik1996autonomous}, or opening drawers \cite{ruhr2012generalized} and doors \cite{doorOpenerSciRob} often require multiple attempts for robots and humans alike. Here, a new ``attempt'' might begin with a re-grasp action to get a better handle on an object, finding the correct pose for the object, such as when inserting a screw driver into a screw, or restarting an activity as its initial conditions have been misjudged, such as during parallel parking. We argue a that trial-and-error approach will remain a persistent feature of autonomy, as one-shot accuracy requires not only perfect perception and information of the environment, but also perfect prediction of the dynamics of the physical world.

\begin{figure}
    \centering
    \includegraphics[width=0.9\textwidth]{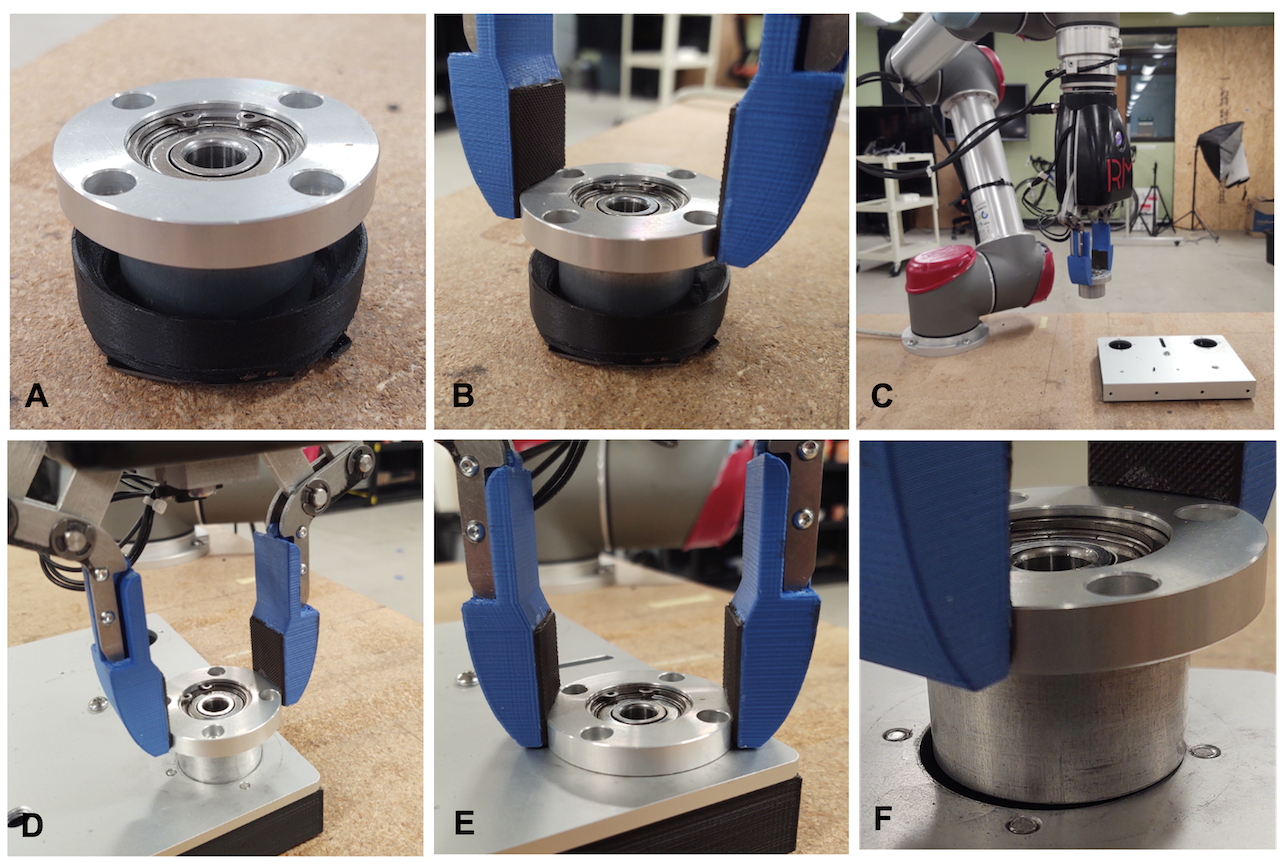}
    \caption{A complete peg-in-hole assembly sequence: \textbf{A} The bearing is presented in a 3D-printed jig, \textbf{B} The bearing is picked up by the robot and transported to the assembly plate \textbf{C}. Force and torque measurements are used to \textbf{D} locate the hole \textbf{E} and complete insertion. Insertion failure due to misalignment \textbf{F}. Friction with the edge of the hole has caused the twisting action to pull the bearing further from the hole center.} 
    \label{fig:setup}
\end{figure}

In this work, we focus on a tight-tolerance insertion problem that we have encountered during a series of industrial robotic competitions \cite{wyk2016robotic,drigalski2018,WRSchallenge}. In our previous work \cite{WRS2018}, we described a series of force-torque (FT) based reactive control schemes for a variety of insertion tasks. We are using one of these tasks, the insertion of a bearing into an assembly, to illustrate the general trial-and-error strategy and how it might be improved. A complete assembly sequence as well as a typical failure condition that requires restart are shown in Figure \ref{fig:setup}. In this task, final success can be detected easily and with low uncertainty, but tight tolerances, non-linearities, and other geometric effects introduce uncertainty that can derail the task and require a restart. Restarting after a failure is one suitable policy, but force-torque data available during task progress provide rich sensory information that enables early detection of failures.  

\begin{figure}
    \centering
    \includegraphics[width=0.9\textwidth]{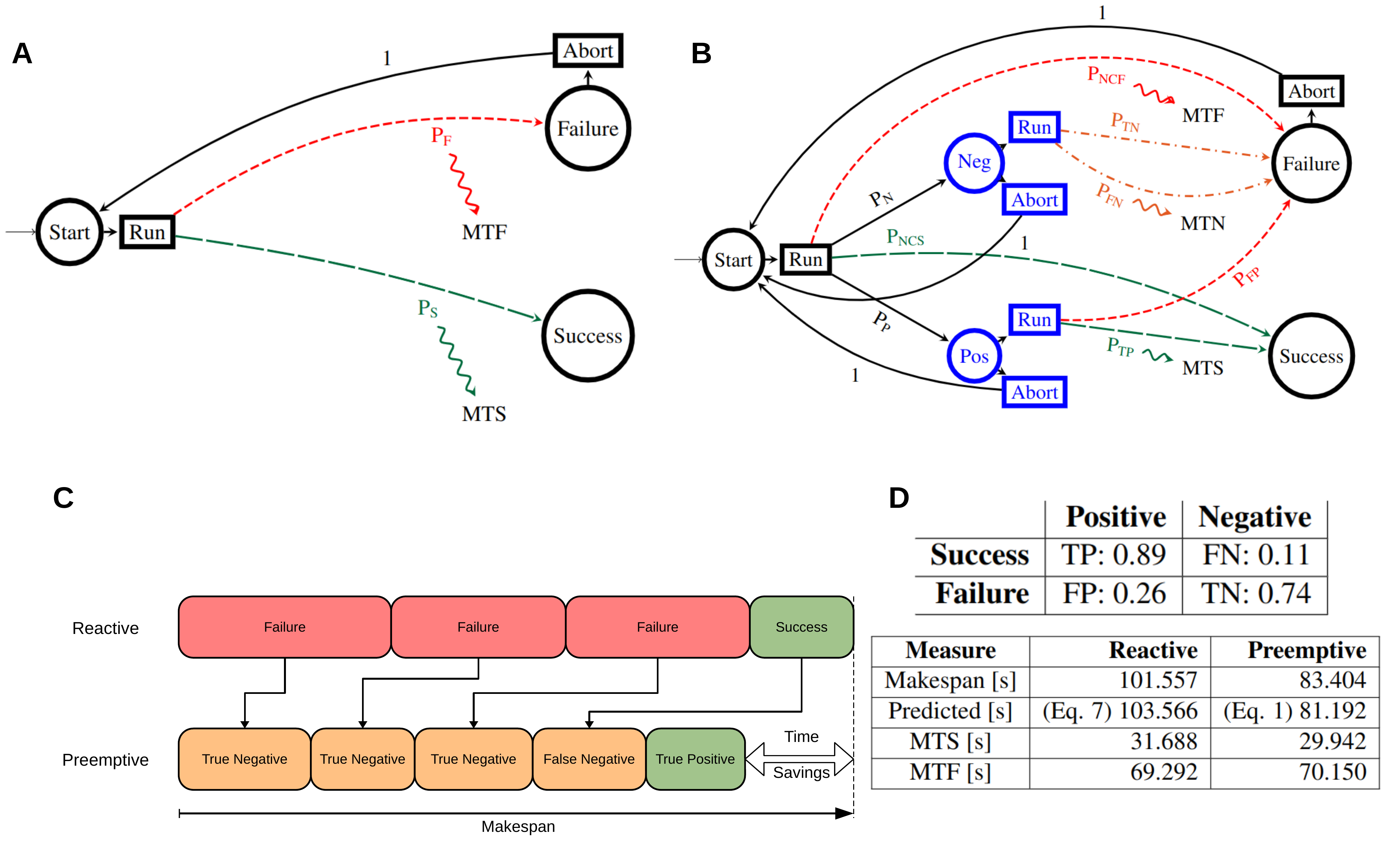}
    \caption{\textbf{A} A Markov Decision Process (MDP) for the trial-and-error problem. Failing at a task (with probability $p_F$) will lead to retry, until success is achieved. Mean-time-to-failure (MTF) and Mean-time-to-success (MTS) are rewards that accumulate to the total makespan. \textbf{B} MDP with predictor, predicting task success (``Pos'') or failure (``Neg'') or succeeding/failing without prior classification. In the above; green, long-dashed paths represent time cost MTS. Red, short-dashed paths represent time cost MTF. Orange, dot-dash paths represent time cost MTN. \textbf{C} Sample task execution with and without error prediction. Failing early may reduce makespan even in the case of false negatives. \textbf{D} mean-time-to-X measurements and confusion matrices for Reactive and Preemptive experiments (N=250).}
    \label{fig:mdp}
\end{figure}

We generalize the class of tasks that are characterized by (1) a low-entropy means to determine terminal success, and (2) a high-entropy means of predicting outcome as a Markov Decision Process (MDP) in which the rewards add up to the total makespan (Figure \ref{fig:mdp}A). 

Specifically, a task failure (with frequency $p_F$) will result in a reward that corresponds to the expected mean-time-to-failure (MTF), whereas a task success (with frequency $p_S$) will result in a reward corresponding to the expected mean-time-to-succeed (MTS).  A complete task consists of zero or more MTF rewards followed by a final MTS reward, leading to the total makespan. We note that the MDP shown in Figure \ref{fig:mdp}A does not leave room for any decisions: restarting in the event of failure is the only suitable policy and will result in the minimum makespan. This is the Reactive scenario. It can only react to a failure signal after it is received.

The addition of a predictor that, after a certain time, predicts a negative (``Neg'') or positive (``Pos'')  outcome, leads to a non-trivial MDP with multiple policy choices (Figure \ref{fig:mdp}B). Assuming the predictor to be always correct, results in a trivial policy --- always abort attempts that will fail. We call this the Preemptive scenario. In the face of time-costly failures, this is of enormous benefit. When the predictor is uncertain, a false negative (FN) might lead to the robot to abort an otherwise successful run. False positives (FP), instead, will not affect the makespan as they will not lead to terminal success. An example sequence is illustrated in Figure \ref{fig:mdp}C. Here, a sequence of three failures is cut short by preemption, leading to an overall savings in makespan, despite a false negative classification and therefore requiring an overall larger number of trials.

Also, there are instances when the classifier itself fails to reach a conclusion (the confidence threshold is not reached) before an attempt ends. These cases must be accounted for when considering the impact of classifier performance on makespan. A non-classified success (NCS), occurring with probability $\p{NCS}$, has no impact on makespan because a predicted success should be allowed to continue to completion.  A non-classified failure (NCF), occurring with probability $\p{NCF}$, represents a missed opportunity to cut a costly failure short. 

\begin{figure}
    \centering
    \includegraphics[width=0.9\textwidth]{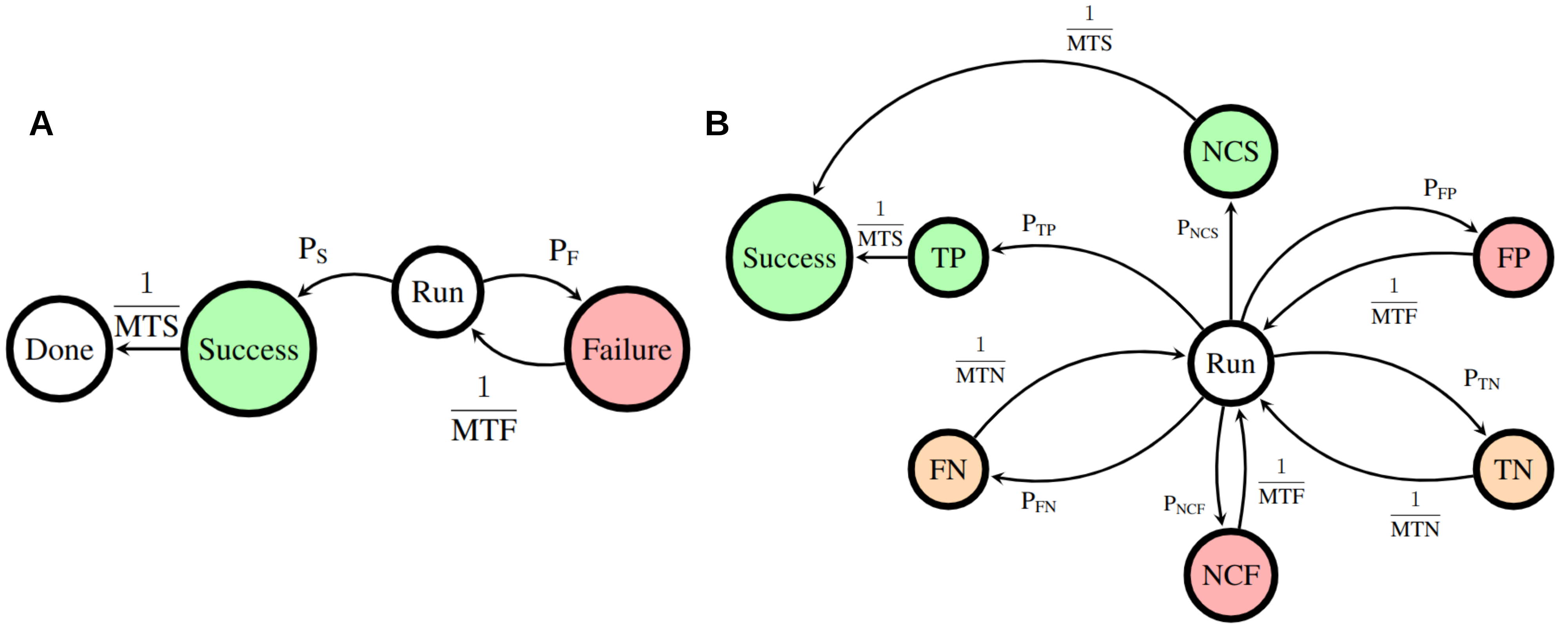}
    \caption{\textbf{A} Markov Jump Process for the MDP in Figure \ref{fig:mdp}A. \textbf{B} Markov Jump Process for the MDP in Figure \ref{fig:mdp}B for a Preemptive policy}
    \label{fig:mjp}
\end{figure}

Finding an optimal policy for an MDP such as shown in Figure \ref{fig:mdp}B, is a classical reinforcement learning \cite{sutton2018reinforcement} problem, with associated transition probabilities and rewards. We note, however, that all the rewards and probabilities can be quantified experimentally and expressed in the form of ``mean-time-to-X'' timing information, the confusion matrix of the classifier, and other frequency-based observations. Assuming that the ``dwell times'' in the relevant states follow a Poisson distribution and adopting a Preemptive policy (abort on negative indication), allows us to reduce the MDP into a Markov Jump process (MJP).  Figure \ref{fig:mjp}A shows the MJP for the basic MDP. Figure \ref{fig:mjp}B shows the MJP for the MDP with task failure prediction and preemption.
Figure \ref{fig:mdp}A and B show the MDPs that govern the Reactive and Preemptive processes. Edges that lead to the termination of an attempt accrue a time cost.  Successful attempts (green, long-dashed) add MTS to the makespan. Failed attempts (red, long-dashed) add MTF to the makespan. Negative classifications (orange, dot-dash) add MTN ($36.26s$) to the makespan. For planning purposes, all other edges have cost zero.

We note that although the ``true'' states are unknown to the robot, corresponding probabilities can be derived from the confusion matrix of the classifier. We can therefore derive a closed-form expression for time saved under a Preemptive policy without having to solve the underlying MDP.

Using an analytical expression for the makespan $t\sub{Run}$, also known as ``sojourn time'', of the Markov Jump Process, we can determine the regime of timing information and classifier attributes in which a Preemptive policy is optimal and when it is not:
\small
\begin{align}
t_{Run}=
\frac{  \left( 1 +
\MTF (\p{FP} + \p{NCF}) + 
\MTS (\p{TP} + \p{NCS}) + 
\MTN (\p{FN} + \p{TN}) 
\right) }{
1- \p{FN} - \p{FP} - \p{TN} 
} \label{closeExpTime} 
\end{align}
\normalsize

A derivation of this equation is provided in the supplemental materials. We note that such an expression is not only of interest during design, but might allow us to switch an to an optimal policy after the robot has undergone sufficient training, i.e.\ acquires a favorable confusion matrix and timing. In the future, a robot might also do this automatically, thereby dramatically increasing its the ability to self-correct and overall autonomy. 

\section{Implementation}

Behavior trees (BT) are a powerful programming abstraction to implement complex reactive behaviors in computer games and robotics. BTs achieve their robustness by monitoring a series of subgoals and reactively starting over until a task is achieved. BTs also require a low-entropy signal for terminal success or failure, while offering the potential to gathering intermediate data that can be used for prediction --- whether it is sufficient for a policy improvement will depend on the confusion matrix of the classifier and overall timing of the process. Due to the suitability of the BT framework for preemption, we introduce the concept of \emph{Preemptive Behavior Trees} (PBT) when combining BTs with an observer that can predict failure based on time series data. 

\begin{figure}
    \centering
    \includegraphics[width=0.9\textwidth]{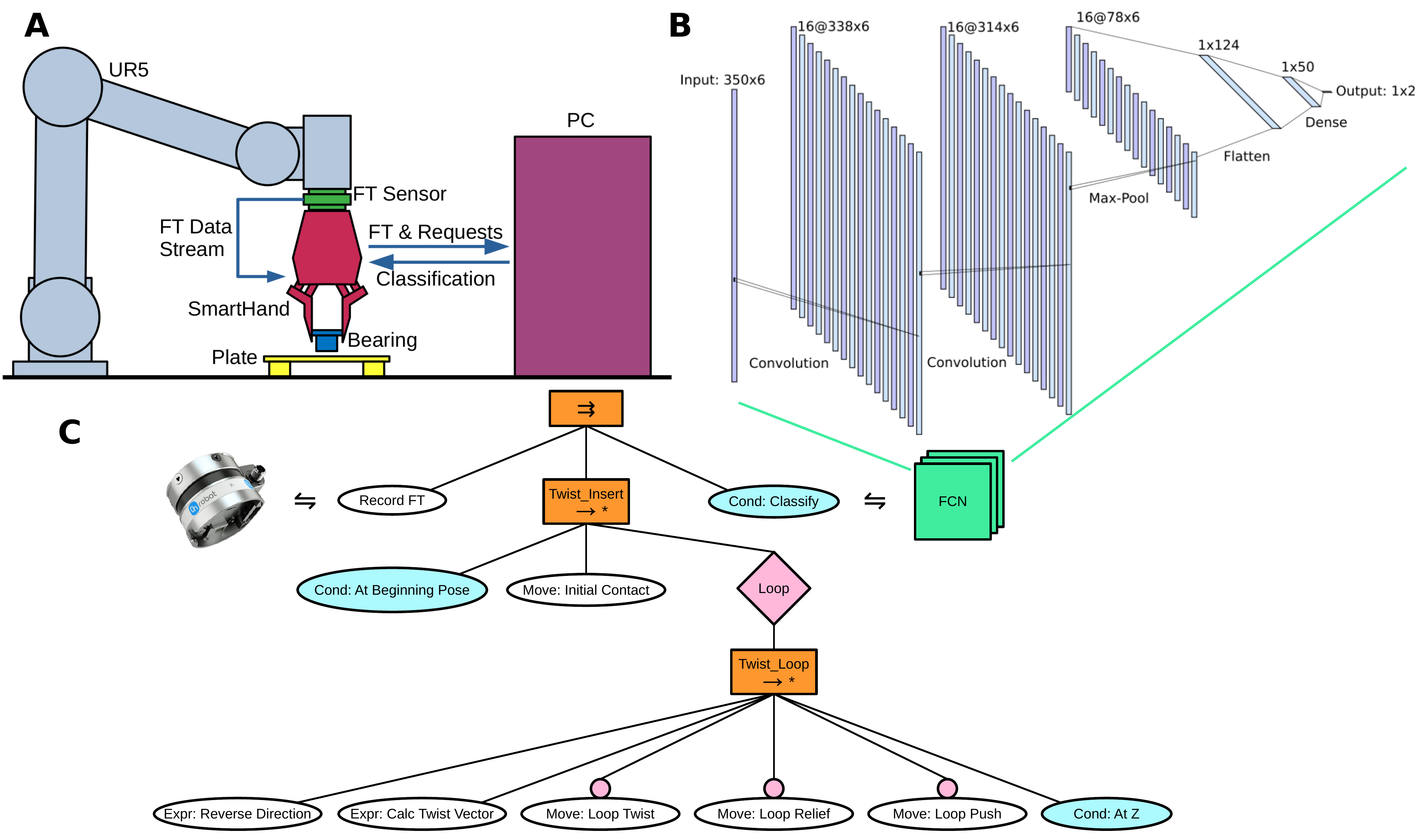}
    \caption{\textbf{A} Robotic setup showing an Universal Robot UR5, an Optoforce F/T sensor, and a Robotic Materials SmartHand. \textbf{B} A dilated convolutional neural network used as a predictor within a Behavior Tree \textbf{C} which maintains a recording process, insertion process, and classification process all in parallel at the top level. The insertion skill will perform the twist insertion sequence in a loop until a successful or 6 failed iterations, whichever comes first.}
    \label{fig:bt}
\end{figure}

Figure \ref{fig:bt}C, shows a PBT skill for the a peg-in-hole assembly task from the World Robot Summit industrial assembly competition using a Universal Robot UR5, an OptoForce Force/Torque sensor, and a Robotic Materials SmartHand \cite{correll2021systems}. The twist insertion skill consists of a sequence that twists the held part, backs off a small amount, then presses down; all in a loop that repeats until it succeeds or times out after N=6 iterations. The physical actions of the insertion behavior are attached to decorators such that they always return success (pink circles). This is because each iteration of the sequence can achieve partial insertion progress, and only at the end of each iteration is the seated status of the bearing is checked by the condition ``At Z.''  As classifier, we are using a dilated convolutional neural network following the approach for time-series classification in \cite{chosenFCN}, which has been pre-trained on 250 task attempts. 

\section{Results} \label{sectResults}


\paragraph{Classifier training using the reactive robot skill} We collected training data for 250 attempts of the twist insertion task.  In order to simulate typical perception errors, the insertion pose was perturbed by $X$ and $Y$ offsets drawn from a normal distribution $\mathscr{N} \sim (0.0, 0.875\text{mm})$. Adding noise allowed us to obtain a sufficient number of failure examples (43.6\%) for training with only 250 attempts. 

\begin{figure}
    \centering
    \includegraphics[width=0.9\textwidth]{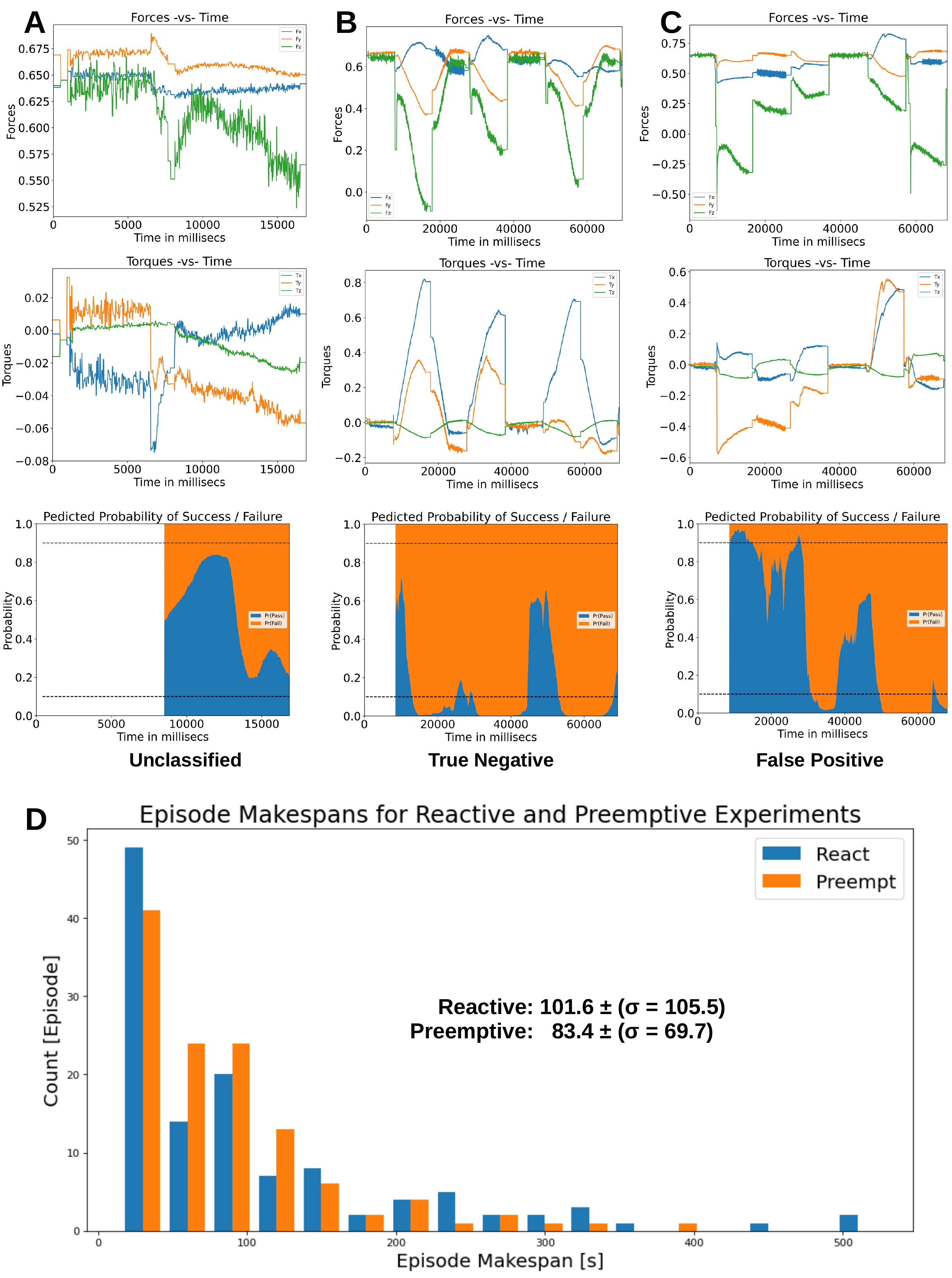}
    \caption{Sample force (top) and torque (middle) readings for experiments that lead to ``unclassified'' (\textbf{A}), true negative (\textbf{B}), and false positive (\textbf{C}) classifications (bottom). (Forces and torques have been normalized.) Prediction only starts after 7500ms. \textbf{D} Histogram showing the makespan distribution for Reactive and Preemptive experiments.} 
    \label{fig:histo}
\end{figure}

The dilated Fully Convolutional Network (FCN) classifier (see Materials \& Methods) was trained over 250 epochs using 350-data point by 6-channel force-torque (FT) samples from the baseline task performance data. In Figures \ref{fig:histo}A-C, the first row has example forces recorded during attempt, with concurrent torques in the second row.  Training data is collected under the Reactive procedure.  Each sample represents 7 seconds of FT data.  The basic dilated FCN model has fairly good characteristics and the ability to identify most failures (Figure \ref{fig:histo}B, bottom row), before the end of an attempt.  The relevant measures of classification and time performance are reported in Figure \ref{fig:mdp}D. 

None of the experiments exhibited catastrophic failure during insertion, such as the work-piece getting stuck, the robot colliding with the environment, or the work-piece slipping out of the robot's grasp.  However, there were some instances when the work-piece slipped from the robot's grasp after insertion, such that human intervention was needed before the next attempt.  These failures occurred after data had already been recorded for the attempt, and are not seen as an influence on the results. 

\paragraph{Experimental validation of preemptive skill efficiency}
In the Reactive case, the average makespan (that is, the accumulated total time on task from first attempt to first success) over 120 episodes is $101.6\pm (\sigma = 105.5)$ seconds. In the Reactive case, about half of all attempts fail, with $p_F = 0.498$ and $p_S = 0.502$. The average makespan over 120 episodes is $83.4 \pm  (\sigma = 69.7)$ seconds for the Preemptive case, for an average time savings of about 18\%. Random position noise was added to the starting pose of all task attempts during the validation experiments. The noise was drawn from a distribution identical to that of the training population: $\mathscr{N} \sim (0.0, 0.875\text{mm})$ in each lateral direction $X$ and $Y$. As the standard deviations of both distributions overlap, we applied the Kruskal-Wallis non-parametric test \cite{SciPy} to confirm that the distributions of makespans are indeed different. To this end, we removed all episodes from both the Reactive and Preemptive experiments that consist of only a single successful attempt, since these episodes represent identical behavior. This yields a p-value of $2.24 \times × 10^{-4}$, so we reject the null hypothesis of Reactive and Preemptive makespans being from the same distribution.

\paragraph{Impact of classifier characteristics on makespan}

\begin{figure}[H]
    \centering
    \includegraphics[width=0.9\textwidth]{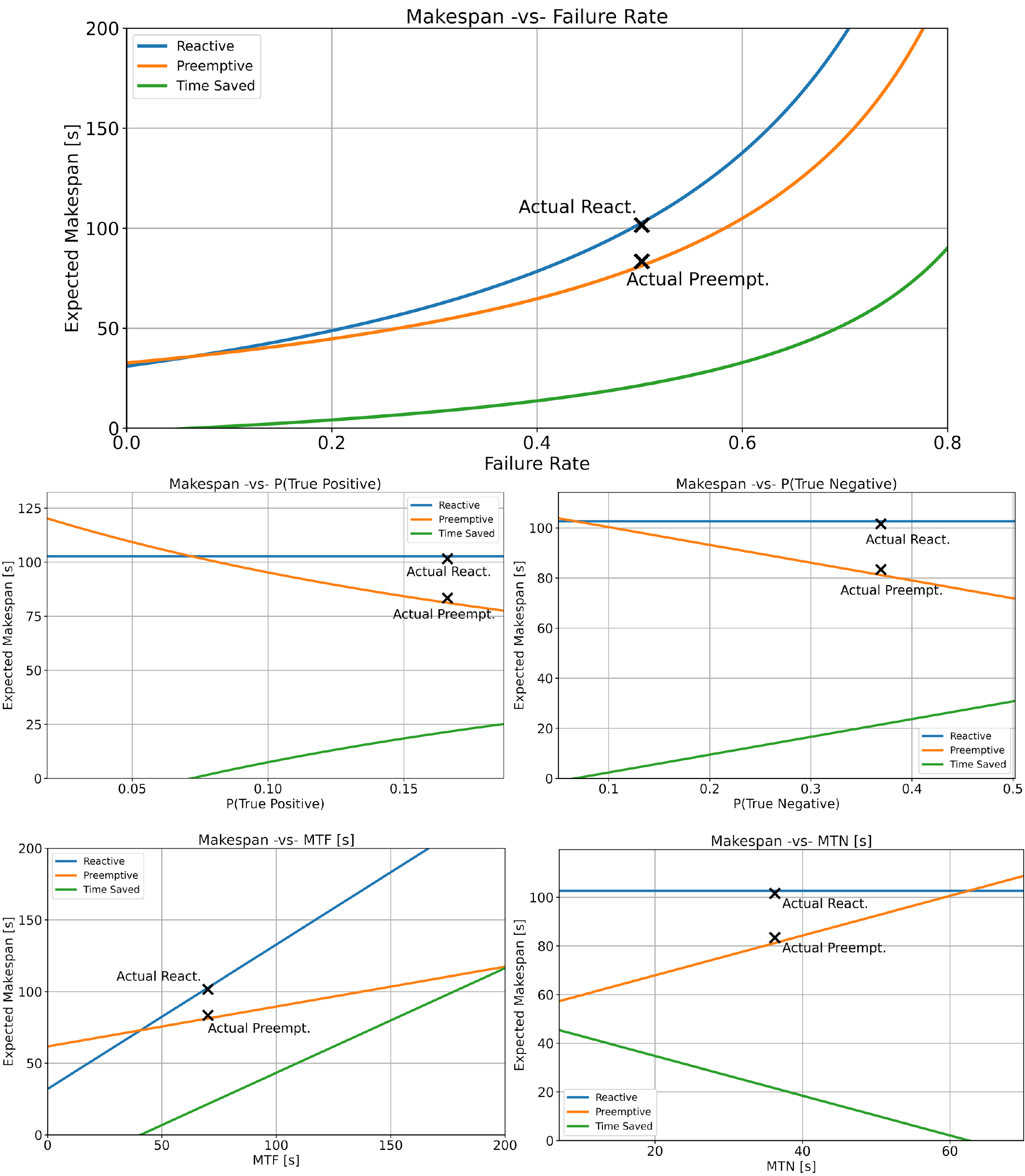}
    \caption{Sweep of success rate and classifier parameters, with time saved.}
    \label{paramSweep}
\end{figure}

Figure \ref{paramSweep} shows the predicted makespan for different combinations of mean-time-to-X and confusion matrix frequencies, indicating regions in which a Preemptive policy is beneficial and where it is not. Data points corresponding to data shown in Figure \ref{fig:mdp}D are highlighted by an ``X''. We observe close alignment between the expected makespan based on our model and experimental data, not exceeding 3\% error. For all plots, Equation \ref{closeExpTime} parameters were taken from experimental data except for the independent variable indicated by each plot.


\section*{Discussion}

We proposed a policy to reduce the makespan in robotic trial-and-error task by interrupting possible failures early. While the advantage of this approach is clear when the chosen predictor exclusively provides true positive and true negative predictions, the occurrence of false negatives  might lead to the interruption of a behavior that would be otherwise successful. For some combinations of confusion matrix entries of the classifier, task timing, and actual success rate, the proposed approach can also increase the makespan (Figure \ref{paramSweep}), motivating an analytic framework to support a decision. 
While success of the preemptive policy also depends on timing, as a general rule, we identify that once the success rate of the trial-and-error behavior approaches one (Figure \ref{paramSweep} shows the corresponding failure rate), or when the true positive rate is very low, the advantages of the proposed method start to diminish. 

We did not plot results for false positives, which are also excluded from Figure \ref{paramSweep}. A false positive is a failing action that is not interrupted by the predictor. In all of our experiments, we assume that the robot can detect whether an assembly has failed or not after the behavior is terminated. In practice, and depending of the task, this might require additional sensors, for example a camera, which again introduces the problem of false positive and negative classifications. We believe the proposed Markov Chain model to easily extend to these cases.

None of the more than two hundred experiments to train the FCN classifier has shown catastrophic failure, defined by failure modes that the robot cannot recover from by itself, and which would invalidate the approach presented here. Such failures might include the work-piece getting jammed, the robot colliding, or the work-piece falling out of the robot's grasp. In order to model these phenomena, the Markov Model shown in Figure \ref{fig:mdp} would need to be extended by an additional absorbing state. As such permanent failures would also affect our baseline method that does not use prediction, we believe using predictors to be still advantageous, even if the system could exhibit catastrophic failure. In future work, we wish to explore whether prediction and early interruption of an experiment that is likely to fail will also help to reduce jamming or losing the work-piece. This will require selecting work-pieces and assembly assignments that will make the occurrence of such events more likely.

As training of the classifier can be done during normal operation of the robot, this provides an opportunity for the robot to decide when to switch to a policy that uses a failure predictor or resume learning. 

\paragraph{Limitations of the study}
The study is limited to scenarios with a clear binary indicator of final success or failure, with noisy in-process monitoring data available at execution time. The study assumes that failures are recoverable, that is, that the option to restart execution will always be available. Catastrophic, unrecoverable failures would change the structure of the jump processes studied by adding a new absorbing state. Sojourn time can still be quantified under a regime that includes catastrophic failures, but Equations \ref{closeExpTime} and \ref{closeRawTime} would no longer apply.

\subsection*{Related Work}

Matsuoka \textit{et al} \cite{valueOfInfo} presents a resilient framework for weighing the value of perception data against the risk of time-costly failures. While \cite{valueOfInfo} considers the cost of additional perception on the makespan, we are developing a general framework to predict the impact of noisy perception on time-efficiency. 
This measure is generally applicable to similar scenarios in which a process is subject to inspection and rework.

Industrial assembly is a well-studied domain concerned with the jigging, sequencing, and physical execution of joining mechanical parts together to form a complete, functional product.  Starting with the first industrial robot arm, Unimate, industrial assembly robots have become faster and more accurate \cite{gasparetto2019unimate}.  More recently, force and vision sensors have found their way into speciality applications in industrial settings \cite{nof2012industrial}.  A good overview about both the state-of-the-art in industrial practice as well as research is provided in \cite{WRSchallenge}. Broadly speaking, current approaches fall into two categories: open-loop approaches that rely on jigging and custom end-effectors, and autonomous ones that use vision and feedback control \cite{WRSchallenge,IKEAsciRob}.

This work is contributing at the intersection of robotic assembly, task planning behavior trees, and machine learning for identifying assembly progress. This work involves a twist insertion assembly behavior, following the taxonomy of \cite{nagele2018prototype}, that relies on in-hand sensing and compliant fingers. 
 
In \cite{nortonBoard} similar behaviors are explored using exclusively commercially available sensing components and end-effectors. In this paper, we are not concerned with the absolute accuracy or reliability of the assembly behaviors, but assume that they are a stochastic process that follows a Poisson distribution, that is, they are characterized by constant rates to either succeed or fail that result in a mean time to succeed or mean time to fail, respectively. 

How to compose the atomic assembly behaviors into complex assembly plans \cite{asmPrimitives} is also a focus of ongoing research. In \cite{asmPrimitives}, the iTASC \cite{iTASC} framework is proposed, but reaches its limitations in automatically sequencing behavior primitives. Using Behavior Trees (BTs) for enhanced robustness of the assembly process has been suggested in \cite{naik2021lessons}, and BTs have been extended to stochastic behavior trees to reflect the different outcomes, and associated behavior therewith, that a robotic experiment can take. In \cite{colledanchise2014performance} a discrete time Markov chain framework has been proposed, but only considers true positive and true negative outcomes. In this paper, we additionally consider scenarios in which a robot chooses inappropriate behavior due to false positive or false negative sensing events.

Planning is strongly related to the specific objectives of the assembly problem at task level. In our previous work \cite{watson2019assembly} we have explored algorithms to find sub-assemblies that maximize the success of an assembly \cite{watson2019assembly}, albeit in a completely deterministic environment.  We consider and minimize the expected error of placement of beams in a truss \cite{komendera2015precise}, and maximizes the stability of the resulting structure \cite{mcevoy2014assembly}. In a domain closely-related to assembly, Adu-Bredu \textit{et al} \cite{elephants} pack groceries while constraining the consideration of high-entropy packing state to pose, while assuming that item labels are low-entropy; resulting in a lower-dimensional belief space and more efficient planning.

While these works plan ahead, \cite{camarinha1996integration} presents an architecture for dispatching, monitoring, and diagnosing assembly action as well as recovering from failures by relying on hand-coded detectors to classify force–torque time-series \cite{lopes1998feature}. Later work uses a support vector machine classifier \cite{forceTraceSVM}, convolutional neural networks \cite{FDICNN} and expert systems \cite{FSMclassify} to monitor assembly progress using similar data. These approaches are on-off classifiers and do not evolve with task progress as the dilated fully-convolutional network (FCN) that we explore in this work, and do not take into account the implications of false positives and negatives. Forecasting methods do exist that evolve over time by adjusting their weights as fresh data comes in \cite{econForecast2013,econForecast2020}, including real-time makespan prediction \cite{petriMakespan}, but these do not consider the evolving quality of the predictor.

\section*{Materials and Methods}


\subsection*{Objectives}

This study is designed to experimentally validate Equations \ref{closeExpTime} and \ref{closeRawTime}, which were derived from statistical principles. We created reactive and preemptive scenarios that match the jump process models presented in Figure \ref{fig:mjp}. We performed a sufficient number of experiments ($N=120$) to show that not only do Equations \ref{closeExpTime} and \ref{closeRawTime} predict the mean makespans for their respective cases, but that a significant ($p<0.05$) and predictable time-savings can be realized with a preemptive policy.

\subsection*{Robotic System}

Control of the system is conducted via a Robotic Materials \emph{SmartHand} \cite{correll2019systems}.  The SmartHand is an integrated computing, vision, and parallel gripper platform designed for use with serial, collaborative \cite{collabSafety} robots.  It integrates an nVidia Jetson TX2 computer for control and image processing.  The computer and Intel RealSense D430 RGB-D camera are mounted to an internal frame that provides passive cooling. The SmartHand is mounted to an OptoForce (now OnRobot) HEX-E 6-axis force-torque (F/T) sensor capable of sensing the wrench at the wrist joint with a resolution of 0.2 Newtons in X-Y and 0.5 Newtons in Z direction for force measurements and 0.010 in X-Y and 0.002 Newton-meters around the Z direction for torque measurements. 

The control software used for this work is a collection of Python 3 libraries for interface with the UR5 robot and SmartHand via Jupyter Notebooks. Here, all sensors and actuators are abstracted into Python libraries that communicate with the sensor and actuator programming interfaces using XML-RPC, Real Time Data Exchange (RTDE), and socket communications.  RTDE is a communication protocol developed by Universal Robots for use with the UR series of cobot manipulators.

\subsection*{Behavior Tree Skills}

Behavior Trees \cite{BTbook} are an execution framework intended to provide more modularity, flexibility, and robustness than the formerly dominant Finite State Machine (FSM).  The Behavior Tree (BT) model is designed to be composable and reactive.  The BT structure is composable because leaf nodes representing individual actions and trees representing complex skills may be used interchangeably and nested to an arbitrary depth.  We are able to reuse actions for multiple tasks.  BTs are reactive because each node flows the status of itself and its children to its parent, all the way to the root.  Therefore, failures are handled at the deepest level of execution possible, and cascade towards the root only as recovery actions at each successive level fail.  The architecture also supports the use of condition nodes paired with a sibling subtree.  The condition node can enforce an invariant, such as ``object in hand'' for a manipulation action, that would signal a task failure if violated.  A condition violation then triggers a recovery skill that either repeats until the condition is met, or triggers a fallback skill at a higher depth.  It is notable that there is a BT equivalent to any FSM or Decision Tree \cite{subsumption}, and we can enjoy the above advantages without sacrificing expressive power.  See Colledanchise and {\"O}gren \cite{BTbook} for a detailed explanation of BT control flow.

We implemented BTs for the IROS2020 Robotic Grasping and Manipulation Competition, built on the \texttt{py\_trees} framework.  All actions and skills performed by our robot during the competition were packaged in nodes and subtrees, from simple leaf node actions, such as moving to a pose, to complex skill subtrees, such as the insertion action described below.  In this work, we focus on the performance of one assembly skill: twist insert.  This skill can perform peg-in-hole in a way that is robust to small errors in estimates of a hole/peg pose.

\subsubsection*{Twist Insert BT Skill} \label{tiltBT}

The twist insert skill is shown in Figure \ref{fig:bt} and is best suited for peg-in-hole operations with a peg diameter of 10mm or larger.  In our experiments twist insert was used to insert a 36mm diameter cylindrical bearing into a similarly-sized hole with a tight fit.  The algorithm proceeds as follows:  Given a hole pose (hand-coded in our work, but easily provided either by vision or by CAD data), the grasped shaft is held vertically above the hole by a preset distance. Then the hand is translated downwards in the global frame until the part makes contact with the hole.  Then, a loop begins that allows $N=6$ failures before itself returning failure.  If an iteration of the insertion sequence completes with a success, then the loop exits early.  Each iteration sequence begins with two Expression nodes that calculate the rotation vector of the twist for this iteration.  The hand rotates a predetermined angle, the direction of which alternates each iteration. The hand lifts a very small distance: 1mm.  We found that the combination of twist and lift (labelled ``Loop Relief'' in Figure \ref{fig:bt}) were more reliable in preventing binding during this task than the insertion skills presented in \cite{WRS2018}. After these two unbinding steps are completed, a final push is executed to seat the bearing in place.  If the Z-component of the hand's final pose is near to the expected value, the BT marks the insertion as a success, and the loop exits with success.  However, if $N$ iterations are completed with a failure status, then the skill returns as a failure.

The Expression Node is our own addition to the BT framework.  It's purpose is to operate on values stored in the global blackboard (dictionary) \cite{BTbook}, thus providing us an outlet to offload state from behavior code, as well as a flexible means to reactively parameterize behaviors.  In short, expression nodes evaluate simple Python expressions in which each \texttt{\$keyname} is a blackboard key whose value is substituted into the expression before evaluation.  The result of the expression is stored under another specified blackboard key, to be retrieved by the relevant behaviors.  We did not explore the metaprogramming capability of Expression nodes, as we had no application for such and doing so would frustrate the normal flow of control of BT execution.

\subsection*{Early Failure Identification} \label{sectEarlyDetect}
As individual snap shots of F/T data are inconclusive, we chose a dilated Fully Convolutional Network (FCN) to classify sequences of F/T data from the insertion behaviors described above (Figure \ref{fig:histo}B-D). 
Specifically, we chose a dilated FCN of the same architecture as Khanna and Narayan \cite{chosenFCN}. Our model differs from \cite{chosenFCN} in the following ways; Dilated Convolution Layer 1 (No. of filters = 16 rather than 8, Filter Size = 4 rather than 2, Dilation Rate = 4 rather than 2), Dilated Convolution Layer 2 (No. of filters = 16 rather than 8, Filter Size = 4 rather than 2, Dilation Rate = 8 rather than 4), and a Fully Connected Layer of 50 units (number of units in this layer not given by \cite{chosenFCN}).  Although Recurrent Neural Networks (RNN), such as GRU and LSTM, have shown admirable performance in failure prediction, \cite{heartPredict,bearingPredict,dataPredict}, simpler FCN have shown similar or better performance \cite{moreira2018online,cui2016multi,parker2020nonlinear}.  Shenfield and Howarth \cite{shenfield2020novel} combined FCN and RNN to exploit the benefits of each in parallel, but our application did not require such an architecture.  A detailed study of network architecture is not part of this work, and only the most basic dilated FCN was used for early failure identification.

The FCN takes an input sample from the F/T stream each timestep; which is a rolling window of 6 channel \{$F_X$, $F_Y$, $F_Z$, $T_X$, $T_Y$, $T_Z$\} readings from the FT sensor located between the gripper and the distal link of the manipulator arm.  The classifier returns a distribution over the classes (Pass, Fail).  With timestep size of $\delta t = 20$ milliseconds (50 Hz, imposed by the constraints of the robotic system), it begins assessing each episode at $t = 7.0$ seconds (the length of the rolling window) after the first contact between the bearing and the bracket.  The classifier reaches a conclusion when confidence in either class exceeds $0.90$.

Classification of an insertion attempt over time seemed to follow 3 different trajectories (Figure \ref{fig:histo}).  Very often, the first window collected at $t = 7.0$ was already assigned a confidence over the $0.90$.  Sometimes confidence would fluctuate several times between classes before settling on the final determination.  Least often, classification trends nearly monotonically towards the final classification.  The sooner the output converges on one classification, the greater the opportunity to avoid time lost to pursuing an action that is likely to fail.

\subsection*{Markov Chain Model of a Task Retry-Loop}

In order to begin to characterize the time impact of early failure prediction, we model a task that either succeeds or fails after some time.  If the task fails, it must be retried again, without limit, until it succeeds.  In the model, a task either immediately succeeds with probability $\p{S}$ or immediately fails with probability $\p{F}$ without our ability to observe it during the trial.  Then, after a Mean Time to Success (MTS) or Mean Time to Failure (MTF), the outcome manifests, and the task is either complete, or execution returns to the ``Run'' state to retry the task.  This is called the Reactive Scenario.  We can imagine such a process governing either a BT that repeats until a condition is met, or a manufacturing process in which a workpiece is sent back for rework until it passes QC inspection. The term Mean Time to Failure, common to industry and process modeling, describes a Poisson process.  The relation
\begin{align}
    \text{MTF} = \int_0^{\infty}\left[ t \lambda e^{-\lambda t} \right] dt = \frac{1}{\lambda}
\end{align}
expresses the probability of a failure occurring at any instant in time, assuming that $\lambda$ is constant.

If we express MTF in units of uniform timesteps (e.g. 1 s), then we can describe the system as a discrete time Markov Chain (DTMC), Figure \ref{fig:mjp}.  The Figure \ref{fig:mjp} process will be called the ``Reactive'' scenario. Failure recovery is mediated by the BT, but there is no possibility for ending a likely failure early. 

We now augment the Reactive model with detection capability.  During the execution of the task, there is an opportunity to predict a failure in progress, and immediately terminate the trial in order to begin the next, thus saving time. In this work, we refer to the task retry loop augmented with early failure identification as the ``Preemptive'' Scenario. There are four possible outcomes: True Positive (TP), a successful assembly gets predicted as such, false negative (FN), a successful assembly gets erroneously classified as failing, true negative (TN), a failing assembly gets classified as such, and false positive (FP), a failing assembly gets erroneously classified as succeeding. The FCN classification process is also assumed to be a Poisson process with a constant probability to predict an outcome after the first 350 data points have been collected.  We model the incidence of the determination made by a failure-prediction scheme by introducing a new term; Mean Time to Negative (MTN).  We note that in a Preemptive policy Mean Time to Positive (MTP) does not meaningfully impact the makespan. We also consider MTN and MTP to be separate quantities, rather than a Mean Time to Classification because, although they are quite similar in this task, they diverged widely for tasks not covered by this work.

Figure \ref{fig:mdp} depicts an absorbing Markov chain model in which ``Done'' is the only absorbing state that can only be achieved from a successful assembly. All jump probabilities are positive nonzero, and the system must evolve towards ``Done'' as $t \rightarrow \infty$ with the stationary distribution $P(\text{Done}) = 1.0$.  Note that job rejection is not modeled.  A failure always returns to ``Run'' for rework until we reach the only acceptable result: ``Done''. 

Any process that has the same steps as Figure \ref{fig:mjp}A, whether manual or automatic, transforms into \ref{fig:mjp}B when in-process monitoring is added.  The simple try-fail-retry process modeled has wide applicability; not only to robotic planning, but in manufacturing as well.  Given two behaviors A and B an autonomous system may choose the shortest running behavior based on how likely it is to catch a failure in progress.  In the industrial setting, the relationships presented here can also aid in a cost-benefit analysis between two process monitoring systems by allowing the purchaser to compare, in concrete terms, the amount of re-work that will be saved.

\subsubsection*{Derivation of Expected Makespan}

Given the parameters for the task \{MTF, MTS\} and the classifier \{MTN, TP, FN, TN, FP\}, we can directly compute the expected running time, or makespan, for this model using the common analysis techniques as described succinctly by Grinstead and Snell \cite{fundMatx} in Theorem 11.5.  In an absorbing Markov process with $\lvert S \rvert$ states, the $\lvert S \rvert \times \lvert S \rvert$ transition matrix $\Pm$ is expressed in canonical form, arranged with the first $M$ rows representing the outgoing edges from the $M$ transient states (Eq. \ref{canon}) followed by the $\lvert S \rvert - M$ absorbing states.  In canonical form, the upper-left $M \times M$ matrix $\Qm$ describes the transitions between the transient states.  The self-loop probabilities of the absorbing states are all $1$ and there are no outgoing transitions, so the lower-right matrix is identity.

Consider a matrix $\Nm$ in which each element $n_{ij}$ is the expected number of times the process is in a transient state $s_j$ before being absorbed after starting in transient state $s_i$.  $\Nm$ is known as the fundamental matrix of transition matrix $\Pm$.  We can find $\Nm$ by computing the infinite series of Equation \ref{infSeries}.  The construction of the infinite series begins with $\Eye$ because the process starts in state $s_i$ by definition, so we must count at least one visit there. Each entry $p_{ij}$ of $\Pm$ is the probability of the process moving from $s_i$ to $s_j$. From the Markov property, the probability of having moved from $s_i$ to $s_j$ after $n$ steps is $p_{ij}$ of $\Pm^n$. So, at each step $k$ we count the transitions by adding $\Qm^k$ to our sum, since we are only concerned about travel between transient states. It is not necessary to carry out this sum, as Theorem 11.4 of \cite{fundMatx} contains a detailed proof that the infinite series has the solution of Equation \ref{fund}.

Given the definition of $\Nm$, we can find the expected sojourn time (expected number of transitions until absorption) starting in each state $s_i$ by summing the elements in row $i$.  This is equivalent to multiplying $\Nm$ by the column vector $\mathbf{c}$, comprised of $M$ elements of value 1. The resultant column vector $\mathbf{t}$ is a column vector in which the $i^\text{th}$ entry is the sojourn time from state $s_i$. In our case, we have modelled the Preemptive Procedure as a Jump Process with (most of the) transition probabilities expressed in terms of average number of seconds spent in each state, so the time to absorption from state $s_1$ (Run) is also the number of seconds we expect the Preemptive Procedure to run until it succeeds (at the ``Done'' state).  The first element of $\mathbf{t}$,  $t_{Run}$, is the desired quantity. (Equation \ref{sojourn})

Note that the outgoing edges from the Run state are not Poisson process jumps, but represent each attempt randomly and ``instantaneously'' entering one of the possible states before idling there until the disposition of the trial is observed.  However, Equation \ref{fund} must include these ``instantaneous'' transitions in the expected sojourn time.  This may partially account for the discrepancy between the running time predicted by Eq. \ref{sojourn} and that observed. It should also be noted that in our experimental setup, the minimum time before either a positive or negative classification must always be 7 seconds, with no possibility of a shorter time. (See ``Classifier training''.) 

\begin{align}
    \Pm^n &= 
    \left[ \begin{matrix}
        \Qm^n & \dotsc \\
        \Mzro & \Eye
    \end{matrix} \right] \label{canon} \\ 
    \Nm &= \Eye + \Qm + \Qm^2 + \cdots \label{infSeries} \\
    \Nm &= \left( \Eye - \Qm \right)^{-1} \label{fund} \\
    \mathbf{t} &= \left[ \begin{matrix}
        t\sub{Run} \\
        \vdots \\
        t_M
    \end{matrix} \right] = \Nm \mathbf{c} \label{sojourn}
\end{align}

We express $t\sub{Run}$ symbolically by substituting the appropriate parameters MTX and $\p{X}$ into the transition probabilities of $\Qm$, as given by Figure \ref{fig:mjp}. After algebraic manipulation,
we arrive at the closed-form solution, Equation \ref{closeExpTime}.

In this work, the term $\p{X}$ is the probability of Event X.  We obtained these quantities by running the classifier on the attempts of the Reactive Procedure.  $\p{X}$ is the count of event X over the total number of attempts.  The possible events are; true positive (TP), false negative (FN), true negative (TN), false positive (FP), non-classified success (NCS), and non-classified failure (NCF). Quantities MTX and $\p{X}$ are computed from individual \textit{attempts}; that is from each run of the tilt insertion BT.  We describe a \textit{trial}/\textit{episode} as a sequence of attempts executed before the task achieves final success. These quantities must be obtained from Reactive experiments, as the Preemptive system must preempt an identified failure and restart.  That is, the monitoring process treats all negative classifications as task failures.  Therefore the actual quantities of TP and FN cannot be known in the Preemptive case. 

Equation \ref{closeExpTime} holds under the following conditions:  
\begin{enumerate}
    \item $\MTN < \MTF$: The (mean) time to identify a failure is less than or equal to the (mean) time it takes an uncaught failure to occur.
    \item $\MTN < \MTS$: The (mean) time to identify a failure is less than to the (mean) time it takes to successfully complete the task.
\end{enumerate}

If $\MTN \geq \MTF$, then MTF is substituted in place of MTN in the appropriate term, and the classifier is unable to save time on average. It can only lose it in False Negative errors.  If $\MTN \geq \MTS$, then FN restart errors are averted, resulting in a different Markov Chain.  If neither of the above conditions are true, then the classifier has no impact on makespan.  Given the performance of the classifier, the above inequalities are reasonable assumptions.

For the sake of comparison, the makespan for the Reactive process shown in Figure \ref{fig:mjp}A, which can be obtained using the same process described by Equations \ref{fund}-\ref{sojourn}, shown in Equation \ref{closeRawTime}.

\begin{align}
T_{makespan}=\frac{ \minus \left( 
\MTF\cd \p{F} \+ \MTS\cd \p{S} \+ 1
\right) }{
\p{F} - 1
}\label{closeRawTime}
\end{align}
where $\p{S}$ and $\p{F}$ are the success and failure rates, per attempt, respectively.




\nocite{*} 

\section*{Acknowledgments}
This work was supported by the National Institute of Standards and Technology (NIST) under grant number 045-FY19-73 (PII) and by the United States Department of Agriculture's (USDA) National Robotics Initiative under grant number 2021-67021-33450. James Watson contributed the main theoretical and experimental results. Nikolaus Correll contributed the theoretical background for Poisson and Jump processes, historical and literature background, and editorial support. Data used in this work for training and verification experiments can be found on GitHub; \texttt{https://github.com/correlllab/Preemptive-BT\_WRS-Bearing-Data}. The authors declare no competing interests.

\newpage

\end{document}